\documentclass{article}
\usepackage{ijcai25}

\usepackage{times}
\usepackage{soul}
\usepackage{url}
\usepackage[utf8]{inputenc}
\usepackage[small]{caption}
\usepackage{graphicx}
\usepackage{amsmath}
\usepackage{amsthm}
\usepackage{booktabs}
\usepackage{algorithm}
\usepackage{algorithmic}
\usepackage[switch]{lineno}

\usepackage{xcolor}
\usepackage{amssymb}
\usepackage{multirow}
\usepackage{booktabs}
\usepackage{tabularx}
\usepackage{subcaption}

\makeatletter
  \let\orig@thanks\thanks
  \newif\if@thanksed
  \@thanksedfalse
  \renewcommand\thanks[1]{%
    \if@thanksed
    \else
      \orig@thanks{#1}%
      \@thanksedtrue
    \fi
  }
\makeatother

\title{MGCA-Net: Multi-Graph Contextual Attention Network for \\ Two-View Correspondence Learning}

\author{
	Shuyuan Lin$^1$\and
	Mengtin Lo$^1$\and
	Haosheng Chen$^2$\footnote{Corresponding Author}\and
    Yanjie Liang$^{3}$\footnote{Corresponding Author}\and
    Qiangqiang Wu$^4$\\
	\affiliations
	$^1$College of Cyber Security, Jinan University, Guangzhou, China\\
	$^2$Chongqing Key Laboratory of Image Cognition, College of Computer Science and Technology, Chongqing University of Posts and Telecommunications, Chongqing, China\\
	$^3$Peng Cheng Laboratory, Shenzhen, China\\
	$^4$Department of Computer Science, City University of Hong Kong, Hong Kong, China\\
	\emails
    swin.shuyuan.lin@gmail.com, 
    ekyp1025@gmail.com, 
    chenhs@cqupt.edu.cn, 
    liangyj@pcl.ac.cn, 
    qiangqwu2-c@my.cityu.edu.hk
}

\begin{document}

\maketitle	

\begin{abstract}
Two-view correspondence learning is a key task in computer vision, which aims to establish reliable matching relationships for applications such as camera pose estimation and 3D reconstruction. However, existing methods have limitations in local geometric modeling and cross-stage information optimization, which make it difficult to accurately capture the geometric constraints of matched pairs and thus reduce the robustness of the model. To address these challenges, we propose a Multi-Graph Contextual Attention Network (MGCA-Net), which consists of a Contextual Geometric Attention (CGA) module and a Cross-Stage Multi-Graph Consensus (CSMGC) module. Specifically, CGA dynamically integrates spatial position and feature information via an adaptive attention mechanism and enhances the capability to capture both local and global geometric relationships. Meanwhile, CSMGC establishes geometric consensus via a cross-stage sparse graph network, ensuring the consistency of geometric information across different stages. Experimental results on two representative YFCC100M and SUN3D datasets show that MGCA-Net significantly outperforms existing SOTA methods in the outlier rejection and camera pose estimation tasks. Source code is available at http://www.linshuyuan.com.
\end{abstract}

\section{Introduction}
\label{sec:intro}
Two-view correspondence is critical for applications like visual localization \cite{chen2024location}, SfM \cite{schonberger2016structure}, SLAM \cite{placed2023survey}, and 3D reconstruction \cite{schmied2023r3d3}. 
By matching features between two images, a reliable geometric relationship is established, which lays the foundation for estimating camera pose and enhancing robust localization in complex scenes.
However, factors such as occlusion, illumination variations, and descriptor inaccuracies often introduce a significant number of outliers (i.e., incorrect matches) \cite{lin2024robust}. These outliers not only reduce matching precision but also propagate errors into downstream tasks, such as camera pose estimation and 3D reconstruction. Consequently, outlier rejection is a critical step to improve the accuracy and robustness of two-view correspondence \cite{brachmann2019neural}.

Traditional outlier rejection methods, such as RANSAC \cite{fischler1981random} and their variants \cite{barath2018graph}, show high robustness and accuracy when dealing with low outlier ratios. 
These methods typically rely on randomly sampling minimal subsets to fit geometric models and validate their validity based on the number of inliers.
However, their performance significantly degrades with increasing outlier ratios, especially in complex geometric scenes or under challenging conditions such as severe illumination and viewpoint changes \cite{ma2021image,lin2022co}.
While traditional methods remain competitive in certain tasks, their efficiency and robustness are limited in high-outlier scenes. Additionally, these methods often depend on user-defined thresholds or specific prior assumptions, restricting their application in real-world environments. 

\begin{figure}[!t]
	\centering
	\includegraphics[width=1\linewidth]{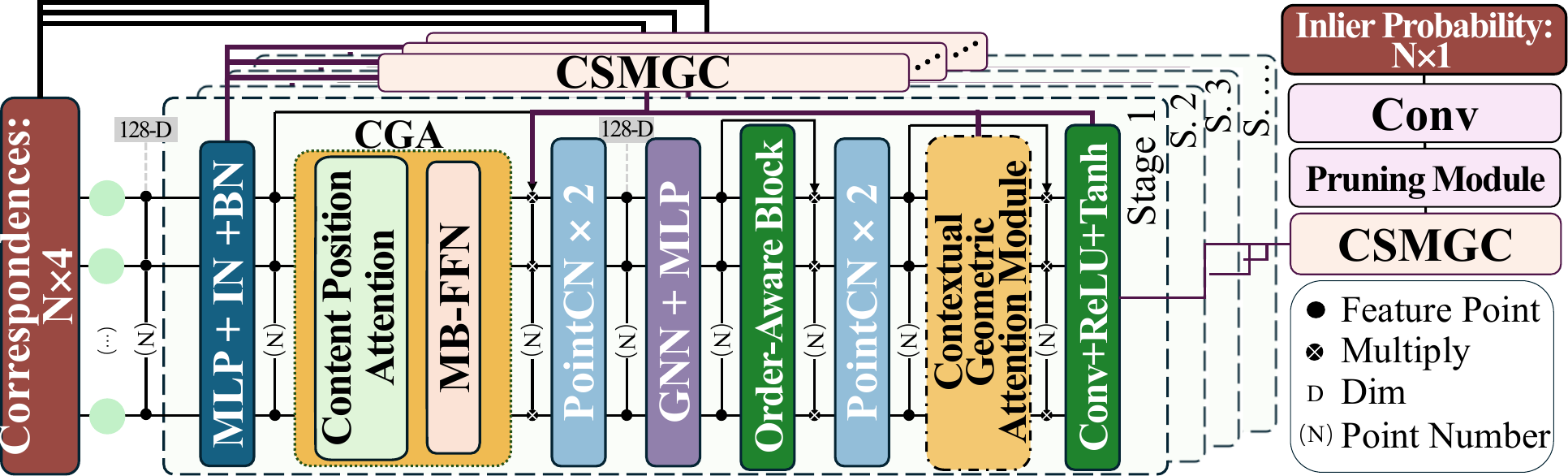}
	\caption{Overall architecture of the proposed MGCA-Net.}
	\label{fig:MGCANet}
\end{figure}

In contrast, deep learning-based outlier rejection methods have demonstrated considerable performance improvements.
PointCN \cite{yi2018learning}, as a pioneering work, modeled the outlier rejection task as a binary classification problem. It utilized multi-layer perceptrons (MLPs) to process unordered keypoints and effectively capture global information through contextual normalization. However, MLPs are inherently limited in capturing local geometric information.

In addition, most deep learning-based methods \cite{zhao2021progressive} rely on simple convolutional operations or fixed neighborhood clustering, which fail to fully capture the complex geometric relationships between keypoints. 
In challenging scenes such as severe illumination changes or significant viewpoint variations, these models struggle to align global semantics with local geometric information, making it difficult to adequately capture and represent the intricate relationships between them.

To overcome these limitations, solutions based on convolutional neural networks (CNNs) and Transformers have been proposed.  For example, ConvMatch \cite{zhang2023convmatch} combines dense motion regularization with local convolutional operations to extract contextual information effectively. VSFormer \cite{liao2024vsformer} fuses visual and geometric features across modalities to enhance representation, while PT-Net \cite{gong2024pt} employs a pyramid Transformer architecture with sparse attention mechanisms to integrate multi-scale motion field information.
Although these methods have improved outlier rejection, most focus solely on processing features from the previous stage while neglecting feature consistency and cross-stage interactions.

To address these challenges, we propose a Multi-Graph Contextual Attention Network (MGCA-Net) , which effectively integrates spatial features and geometric consensus to enhance outlier rejection. As illustrated in Fig. \ref{fig:MGCANet}, MGCA-Net consists of two core modules: Contextual Geometric Attention (CGA) module and Cross-Stage Multi-Graph Consensus (CSMGC) module, designed to address the limitations of existing methods in local geometric modeling and cross-stage information optimization.
Specifically, CGA consists of a Context Position Attention (CPA) and a Multi-Branch Feed Forward Network (MB-FFN). CPA dynamically fuses spatial and contextual information to effectively balance global semantics and local geometric details, while MB-FFN integrates multi-scale features to enhance feature representation in complex scenes. 
Furthermore, CSMGC uses a cross-stage sparse graph neural network to establish geometric consensus across different stages, enhancing feature interaction and ensuring geometric consistency throughout the process. The main contributions of this paper are summarized as follows:

\begin{itemize}
	\item We propose a novel MGCA-Net that integrates global and local information with multi-stage feature fusion and geometric modeling, enhancing robustness and representation in high-outlier scenes.
    \item We propose CGA, which comprises CPA and MB-FFN. CPA effectively balances global semantics and local geometry by combining spatial and contextual information, while MB-FFN enhances feature representation in complex scenes by integrating multi-scale features.
    \item We propose CSMGC, which establishes geometric consistency across stages by incorporating geometric priors with multi-sparse graph neural networks, significantly improving robustness in outlier rejection.
\end{itemize}
By fusing cross-stage information, MGCA-Net can accurately recognize outliers and effectively reject them even in highly outlier scenes, while dynamically refining correspondence reliability through progressive consensus learning.

\section{Related Work}
\label{sec:RW}
Outlier rejection in two-view correspondence tasks is a key research topic in computer vision.
Various methods have been proposed to improve correspondence accuracy, which can be broadly categorized into three main types: traditional methods, learning-based methods, and attention-based methods.

\subsection{Traditional Methods}
Traditional methods play a crucial role in two-view correspondence learning, particularly in the outlier rejection and geometric consistency modeling. These methods often rely on hypothesis-validation strategies, with RANSAC (Random Sample Consensus) \cite{fischler1981random} being the most classic and widely applied approach. RANSAC iteratively samples minimal subsets of data, fits geometric models, and evaluates the number of inliers to identify the best model. However, RANSAC is inefficient as it typically requires a large number of iterations to produce meaningful results. 
To address this, USAC (Universal Sample Consensus) \cite{raguram2012usac} integrates several RANSAC enhancements, including dynamic sampling, hypothesis test optimization, and model validation strategies, providing a unified sampling framework.
In addition to hypothesis-validation methods, non-parametric models, such as VFC \cite{ma2014robust}, distinguish inliers from outliers by establishing sparse vector field models and improve adaptability to complex scenes through regularization constraints. 
LPM (Locality Preserving Matching) \cite{ma2019locality} introduces local consistency constraints to efficiently eliminate erroneous matches with linear-logarithmic complexity.
Despite significant advancements in outlier rejection, traditional methods still face limitations in handling high outlier ratios and complex geometric scenes.

\subsection{Learning-Based Methods}
To effectively address the robustness challenges posed by high outlier ratios and complex geometric scenes, deep learning approaches have become a primary solution for outlier rejection tasks. 
LFGC-Net \cite{yi2018learning} is a pioneering work that introduces a simple yet effective PointCN-like structure, reformulating the feature point matching task as an inlier classification problem. However, the single-stage network architecture adopted by LFGC-Net fails to effectively leverage local contextual information, resulting in suboptimal performance in high-outlier scenes.
To overcome these limitations, some methods have introduced iterative networks and pruning strategies to alleviate class imbalance while capturing geometric information.
For instance, MSA-Net \cite{zheng2022msa} integrates multi-scale attention mechanisms into a multi-stage network to improve robustness, and MS2DG-Net \cite{dai2022ms2dg} uses a sparse semantic dynamic graph to dynamically update matching features, enhancing the semantic consistency. 
Additionally, GCA-Net \cite{guo2023learning} and SGA-Net \cite{liao2023sga} use graph attention mechanisms to effectively combine local and global information, improving accuracy and robustness. 
Following this, NCMNet \cite{liu2024ncmnet} further incorporates a neighborhood consistency mining module, capturing geometric relationships between local and global neighborhoods via a sparse graph structure, optimizing matching performance in noisy environments. 

However, they still rely on local consistency scores and lack a deep understanding of global geometric information. Furthermore, during outlier filtering, pruning strategies may incorrectly eliminate inliers, especially in high-outlier scenes, which significantly reduces the number of remaining inliers.
To address these challenges, we propose CSMGC, which dynamically integrates geometric information across stages and enhances feature consistency modeling.

\subsection{Attention-Based Methods}
Attention mechanisms have become a crucial component in deep learning models, particularly in computer vision tasks, where they play an essential role in enhancing models' performance. 
In the task of two-view correspondence learning, attention mechanisms are commonly used for feature extraction. For instance, SuperGlue \cite{sarlin2020superglue} combines self-attention and cross-attention, where self-attention focuses on the representation of features in a single image, while cross-attention compares feature similarities between different images to improve matching accuracy. LoFTR \cite{sun2021loftr} employs a detector-free strategy, using attention mechanisms to capture complex geometric information and generate high-quality matching results. 
T-Net \cite{zhong2021t} employs a Permutation-Equivariant Context Squeeze-and-Excitation module, dynamically adjusting feature weights using channel attention to strengthen global context modeling. 
However, existing attention-based methods face limitations in balancing local and global features, fusing multi-scale information, and handling complex geometric structures, which limits their performance in challenging scenes.
To address these issues, we propose CGA, which combines position encoding with attention mechanisms to preserve spatial relationships between features, thereby enhancing the capture of local geometric features. 

\section{Methodology}
\subsection{Problem Formulation}
\label{sec:problem_formulation}
Given a pair of images $I_1$ and $I_2$ from the same scene, feature points and descriptors are initially extracted from both images using existing methods (e.g., SIFT \cite{lowe2004distinctive}, SuperPoint \cite{detone2018superpoint}). Then, an initial set of correspondences  $S = {s_1, s_2, ..., s_N} \in \mathbb{R}^{N \times 4}$ is generated using a nearest-neighbor matching strategy, where $N$ is the number of initial correspondences. Each correspondence  $s_i = (x_i, y_i, x'_i, y'_i)$ consists of the coordinates of a keypoint in $I_1$ and its corresponding point in $I_2$. Finally, the network outputs logit values and the initial correspondence set $S$, which are input into the eight-point algorithm \( g \), to estimate the fundamental matrix $\hat{E}$.
\subsection{Contextual Geometric Attention Module}
\label{sec:cgam}

\begin{figure}
	\centering
    \includegraphics[width=1\linewidth]{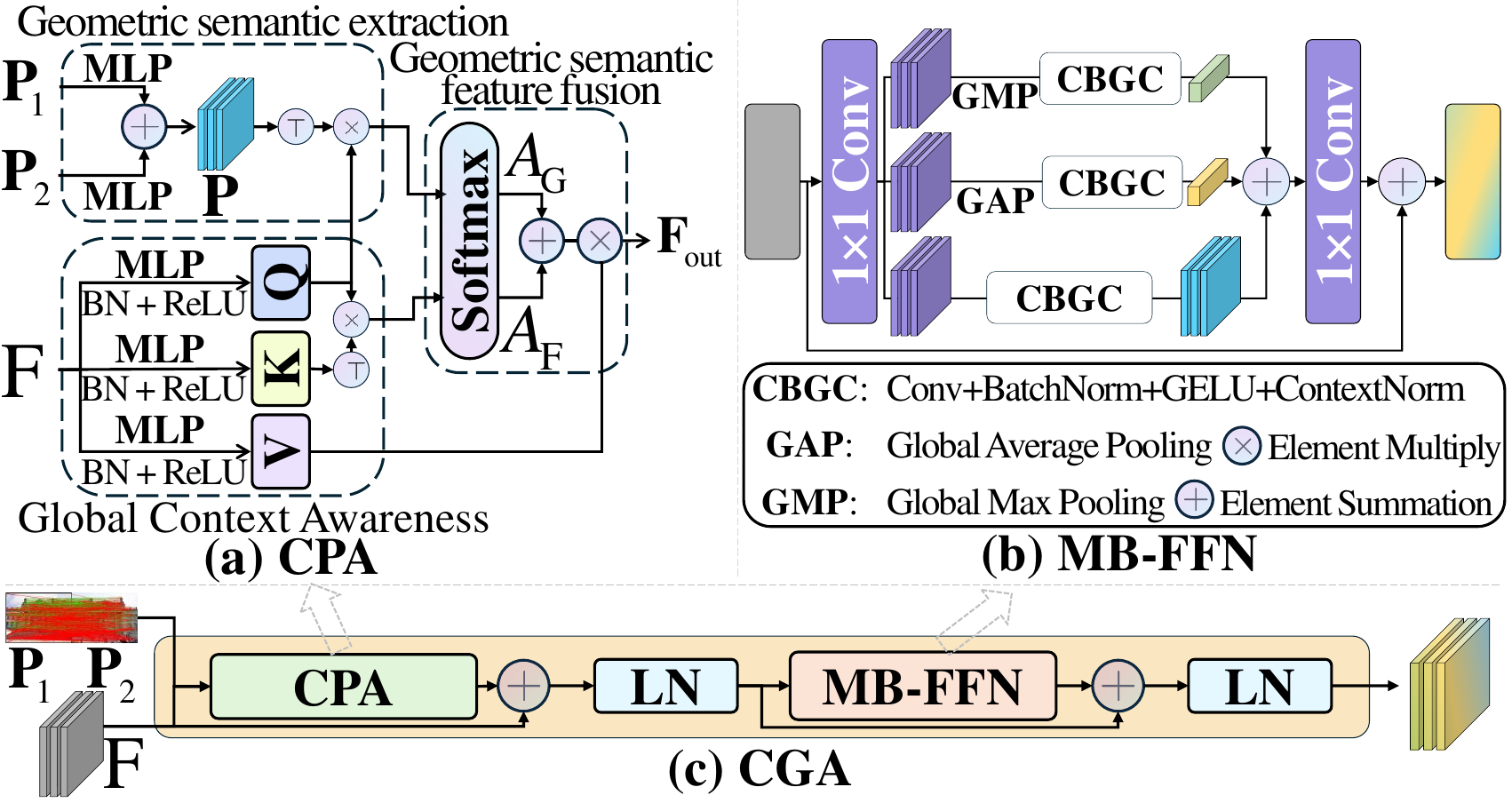}
    \caption{Pipeline of CGA.}
	\label{fig:CGAM}
\end{figure}

As traditional feature extraction methods have difficulty in representing both global semantics and local geometric features, we propose the CGA Module, which enhances feature representation by leveraging spatial and contextual information. As illustrated in Fig.~\ref{fig:CGAM}, CGA consists of two key components: CPA and MB-FFN.

\subsubsection{3.2.1 Context Position Attention}

%

CPA aims to improve the accuracy and robustness of feature point matching in two-view tasks by fusing global context and local geometric relationships. To achieve this, CPA incorporates a dual-attention mechanism (i.e., content attention and positional attention) and achieves dynamic integration of global semantics and local geometric information through the collaborative functioning of three key components: Global Context Awareness (GCA), Geometric Semantic Extraction (GSE), and Geometric Semantic Feature Fusion (GSFF).

\textbf{1) Global Context Awareness}: To model the geometric relationships among global features, CPA processes the input features \(\mathbf{F} \in \mathbb{R}^{N \times d}\) through a Multi-Layer Perceptron (MLP), Batch Normalization (BN), and the non-linear activation function ReLU, mapping the features into three separate spaces, query (\(\mathbf{Q}\)), key (\(\mathbf{K}\)) and value (\(\mathbf{V}\)), as follows:
\begin{equation} 
\begin{split}
	\mathbf{Q} &= \text{ReLU}(\text{BN}(\text{MLP}(\mathbf{F}))), \\
    \mathbf{K} &= \text{ReLU}(\text{BN}(\text{MLP}(\mathbf{F}))), \\
    \mathbf{V} &= \text{ReLU}(\text{BN}(\text{MLP}(\mathbf{F}))).
\end{split}
\end{equation}

The correlation matrix is then calculated as:
\begin{equation}
	A_{\text{F}} = \text{Softmax}\left( \frac{\mathbf{Q} \mathbf{K}^T}{\sqrt{d}} \right),
\end{equation}
where \(\sqrt{d}\) is a scaling factor that balances the numerical range of the attention scores ensuring training stability. The Softmax function normalizes the correlation matrix to produce attention weights, effectively capturing long-range dependencies among feature points.

\textbf{2) Geometric Semantic Extraction}: To address the limitations of existing methods in modeling local geometric relationships, CPA explicitly captures pairwise geometric relationships through a positional attention mechanism. The input feature point coordinates \(\mathbf{P}_1 , \mathbf{P}_2 \in \mathbb{R}^{N \times 2}\) are mapped into a high-dimensional space and encoded into geometric features 
\(\mathbf{P}\) via MLP. Subsequently, the geometric features \(\mathbf{P}\) are combined with the query features \(\mathbf{Q}\) to achieve deep fusion of geometric and semantic information, enabling effective modeling of geometric-semantic relationships between feature pairs. Specifically, the calculation for geometry-enhanced attention is as follows:
\begin{equation}
	\mathbf{P} = \text{MLP}(\mathbf{P}_1) + \text{MLP}(\mathbf{P}_2),
\end{equation}
\begin{equation}
	\mathbf{A}_{\text{G}} = Softmax\left( \mathbf{Q} \mathbf{P}^T \right).
\end{equation}
This component ensures an explicit and robust encoding of pairwise geometric relationships while integrating them with semantic information.

\textbf{3) Geometric Semantic Feature Fusion}: To address the separation of geometric semantics and global features in different feature spaces, CPA introduces a fusion mechanism. 
Specifically, CPA achieves deep integration of geometric features and semantic information by combining the geometry-enhanced attention weights \(\mathbf{A}_{\text{G}}\) and the global context attention weights \(\mathbf{A}_{\text{F}}\). The fused output are calculated as follows:
\begin{equation}
	\mathbf{F}_{\text{out}} = (\mathbf{A}_{\text{G}} + \mathbf{A}_{\text{F}}) \mathbf{V},
\end{equation}
where \(\mathbf{A}_{\text{G}}\) represents the geometry-enhanced attention, \(\mathbf{A}_{\text{F}}\)  denotes the global context attention, and \(\mathbf{V}\) is the value feature. The fused output 
\(\mathbf{F}_{\text{out}}\) integrates geometric and semantic information, encapsulating both the geometric constraints between feature pairs and the dependencies on global contextual information.
Based on the above collaboration, CPA achieves deep fusion of geometric and semantic information within the feature space, significantly enhancing the representational capacity and flexibility of MGCA-Net across features of varying scales.

\subsubsection{3.2.2 Multi-Branch Feed Forward Network}
Traditional Feedforward Neural Networks (FFNs) can perform nonlinear transformations on input features but they struggle to effectively capture both global and local information under complex geometric contexts. To effectively integrate multi-scale feature information and enhance the generalization capability of CPA, we propose MB-FFN. MB-FFN introduces a multi-branch structure that processes features in parallel through different branches, fusing them in the final stage, fully extracting the semantic information of multi-scale features. As illustrated in Fig.~\ref{fig:CGAM} (b), MB-FFN comprises three components: 1) Local Convolutional Branch; 2) Global Average Pooling Branch; and 3) Max Pooling Branch. The outputs from these branches are transformed through the Convolution-BatchNorm-GELU-Convolution (CBGC) module and fused using a summation operation. The final output \(\mathcal{H}\) is calculated as:
\begin{equation}
	\begin{split}
		\mathcal{H} = &\ CBGC(GAP(\mathbf{\text{LN}(\mathbf{F}_{\text{out}}}))) \\
		&+ CBGC(GMP(\mathbf{\text{LN}(\mathbf{F}_{\text{out}}}))) \\
		&+ CBGC(\mathbf{\text{LN}(\mathbf{F}_{\text{out}}})),
	\end{split}
\end{equation}
where $\text{LN}$ represents Layer Normalization, $ CBGC = Conv(GELU(BN(Conv(x))))$, and $GAP$ represents Global Average Pooling, and 
$GMP$ represents Global Max Pooling. These features are progressively transformed through the CBGC module and fused into the output \(\mathcal{H}\) via addition. This design effectively captures global and local features from different perspectives and enhances the model's ability to represent multi-scale features in complex scenes.

\subsection{Cross-Stage Multi-Graph Consensus Module}
\label{sec:CSMGC}
In order to enhance the local geometric constraints and global information fusion between stages, we propose the CSMGC module, which consists of three components: 1) cross-stage feature extraction, 2) graph consensus construction, and 3) cross-stage graph consensus aggregation.

\textbf{1) Cross-stage Feature Extraction}:
To effectively capture geometric features and contextual semantic information from the previous and cross stages, we extract three key features \( \mathbf{Z}_1^{(M-1)}, \mathbf{Z}_2^{(M-1)}, \mathbf{Z}_3^{(M-1)} \) from different modules to represent the feature distribution of the first stage network. Among them, \( \mathbf{Z}_1^{(M-1)} \) refers to features extracted by the first CPA module of the previous stage, representing the global semantic relationship; \( \mathbf{Z}_2^{(M-1)} \) refers to the processing obtained by the second CPA module after PointCN, OANet and PointCN, representing richer local and global information; \( \mathbf{Z}_3^{(M-1)} \) refers to the final combination of MLP and residual information, representing the comprehensive features of the current stage. In addition, to capture the geometric consistency across stages, we perform feature extraction on the network outputs across stages \(\mathbf{Z}_{3}^{(1)}, \mathbf{Z}_{3}^{(2)}, \dots, \mathbf{Z}_{3}^{(M-2)}\), which can be expressed as:
\begin{equation}
	\mathbf{Z} = SE\left(concat\left(\mathbf{Z}_{3}^{(1)}, \mathbf{Z}_{3}^{(2)}, \dots, \mathbf{Z}_{3}^{(M-2)}\right)\right),
\end{equation}
where \(\mathbf{Z}_{3}^{(1)}, \mathbf{Z}_{3}^{(2)}, \dots, \mathbf{Z}_{3}^{(M-2)}\) represent the feature representations extracted from stages \(1\) to \(M-2\) across stages, respectively. These features are concatenated through the \(concat(\cdot)\) operation to form a global representation containing multi-stage features. Subsequently, the fusion module \(SE(\cdot)\) adjusts the weights and compresses the concatenated features to dynamically model the relative importance of the features, thereby generating the final fused feature \(\mathbf{Z}\). 

\begin{figure}
	\centering
	\includegraphics[width=1\linewidth]{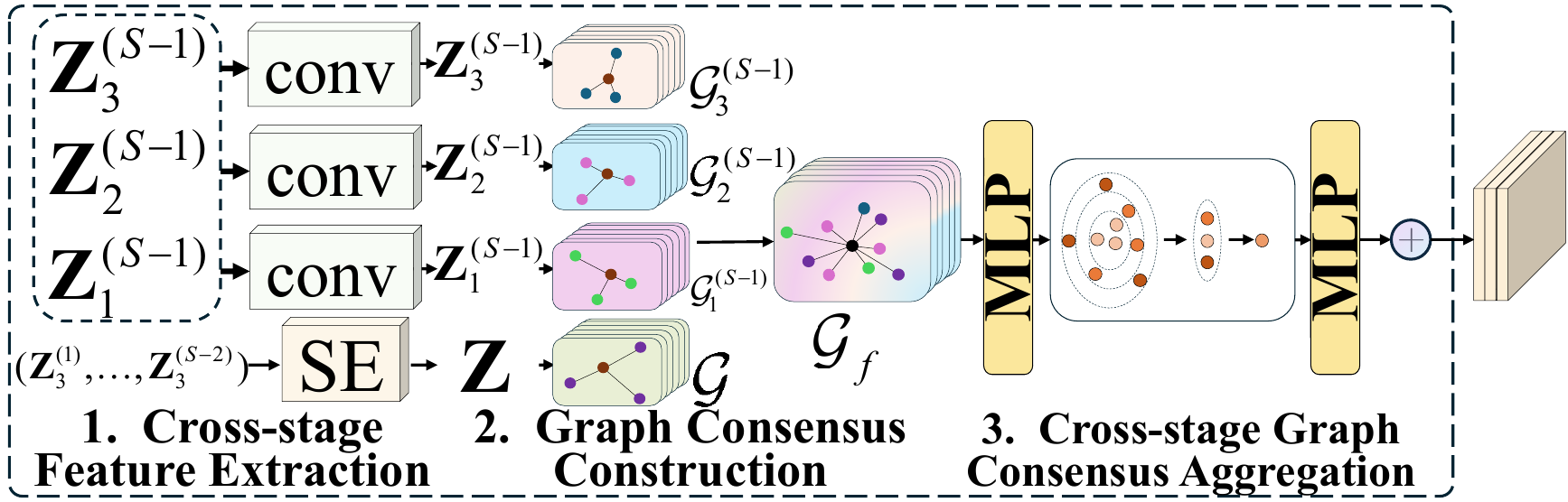}
	\caption{Overall architecture of the proposed CSMGC.}
	\label{fig:CSMGC}
\end{figure}

\begin{table*}[!htbp]
    \centering
    \resizebox{0.8\linewidth}{!}{ 
        \begin{tabular}{c|ccc|ccc|ccc|ccc}
            \toprule
            Dataset & \multicolumn{6}{c}{YFCC100M  (\%)}            & \multicolumn{6}{c}{SUN3D  (\%)} \\
            \midrule
            \multirow{2}[4]{*}{Method} & \multicolumn{3}{c|}{Known Scene} & \multicolumn{3}{c|}{Unknown Scene} & \multicolumn{3}{c}{Known Scene} & \multicolumn{3}{c}{Unknown Scene} \\
            \cmidrule{2-13}          & \multicolumn{1}{c|}{P  (\%)} & \multicolumn{1}{c|}{R  (\%)} & F  (\%) & \multicolumn{1}{c|}{P  (\%)} & \multicolumn{1}{c|}{R  (\%)} & F  (\%) & \multicolumn{1}{c|}{P  (\%)} & \multicolumn{1}{c|}{R  (\%)} & F  (\%) & \multicolumn{1}{c|}{P  (\%)} & \multicolumn{1}{c|}{R  (\%)} & F  (\%) \\
            \midrule
            RANSAC \cite{fischler1981random} & 47.35 & 52.39 & 49.47 & 43.55 & 50.65 & 46.83 & 51.87 & 56.27 & 53.98 & 44.87 & 48.82 & 46.76 \\
            PointNet++ \cite{qi2017pointnet++} & 49.62 & 86.19 & 62.98 & 46.39 & 84.17 & 59.81 & 52.89 & 86.25 & 65.57 & 46.30 & 82.72 & 59.37 \\
            LFGC-Net \cite{yi2018learning} & 54.43 & 86.88 & 66.93 & 52.84 & 85.68 & 65.37 & 53.70 & 87.03 & 66.42 & 46.11 & 83.92 & 59.52 \\
            DFE-Net \cite{ranftl2018deep} & 56.72 & 87.16 & 68.72 & 54.00 & 85.56 & 66.21 & 53.96 & 87.23 & 66.68 & 46.18 & 84.01 & 59.60 \\
            ACNe-Net \cite{sun2020acne} & 60.02 & 88.99 & 71.69 & 55.62 & 85.47 & 67.39 & 54.11 & 88.46 & 67.15 & 46.16 & 84.01 & 59.58 \\
            \midrule
            OANet \cite{zhang2019learning} & 61.14 & 88.16 & 69.73 & 57.90 & 85.07 & 66.53 & 54.43 & 88.08 & 63.72 & 46.50 & 83.83 & 56.32 \\
            T-Net \cite{zhong2021t} & 61.18 & 89.94 & 70.47 & 57.18 & 87.01 & 66.73 & 55.01 & 88.36 & 64.18 & 46.50 & 83.98 & 56.33 \\
            PESA-Net \cite{zhong2022pesa} & 61.43 & 89.63 & 72.90 & 58.02 & 87.01 & 69.62 & 55.08 & 88.56 & 67.92 & 47.29 & 84.81 & 60.72 \\
            MSA-Net \cite{zheng2022msa} & 59.27 & 90.28 & 68.92 & 56.49 & 88.60 & 66.46 & 56.09 & 87.57 & 64.71 & 48.64 & 83.81 & 57.89 \\
            MS²DG-Net \cite{dai2022ms2dg} & 64.24 & 89.31 & 72.49 & 60.38 & 86.71 & 68.96 & 55.58 & 89.01 & 64.63 & 47.42 & 84.50 & 57.12 \\
            U-Match \cite{li2023u} & 63.29 & \underline{92.12} & 72.56 & 61.02 & \underline{90.64} & 70.59 & 55.29 & \textbf{89.35} & 64.53 & 47.69 & \textbf{85.6} & 57.53 \\
            PGFNet \cite{liu2023pgfnet} & 59.21 & 90.22 & 68.77 & 56.38 & 88.15 & 66.20 & 55.80 & 87.80 & 64.68 & 48.37 & 84.03 & 57.89 \\
            GCA-Net \cite{guo2023learning} & 63.82 & 91.36 & 72.87 & 60.44 & 88.92 & 69.74 & 55.00 & 89.08 & 64.21 & 46.88 & 85.14 & 56.79 \\
            ConvMatch \cite{zhang2023convmatch} & 63.14 & 91.2  & 72.21 & 60.22 & 89.48 & 69.65 & 55.79 & 89.23 & 64.89 & 48.13 & \underline{85.55} & 57.87 \\
            ConvMatch \textsuperscript{+}\cite{zhang2023convmatch_plus} & 62.75 & 92.05 & 72.11 & 59.41 & 90.12 & 69.13 & 55.51 & \underline{89.30} & 64.80 & 47.60 & 85.51 & 57.53 \\
            NCM-Net \cite{liu2024ncmnet} & 77.92 & 81.41 & 79.25 & 76.83 & 78.61 & 77.45 & 66.15 & 74.59 & 69.38 & 60.92 & 68.94 & 64.04 \\
            PT-Net \cite{gong2024pt} & 65.69 & 90.61 & 74.14 & 62.14 & 89.22 & 71.16 & 55.41 & 89.17 & 64.66 & 47.45 & 85.52 & 57.39 \\
            DeMatch \cite{zhang2024dematch} & 61.33 & \textbf{92.77} & 71.19 & 58.74 & \textbf{91.02} & 68.91 & 56.00 & 88.90 & 65.10 & 48.27 & 85.24 & 58.07 \\
            BCLNet \cite{miao2024bclnet} & \underline{78.36} & 82.23 & 79.87 & \underline{77.90} & 80.07 & \underline{78.73} & 66.20 & 74.12 & 69.19 & 61.14 & 68.33 & 63.92 \\
            MSGSA \cite{lin2024multi} & 63.29 & 91.37 & 72.40 & 60.03 & 89.34 & 69.53 & 55.92 & 88.56 & 65.08 & 47.99 & 84.53 & 57.81 \\
            CGR-Net \cite{yang2024cgr} & 78.31 & 82.34 & \underline{79.90} & 77.22 & 79.61 & 78.14 & \underline{66.46} & 74.46 & \underline{69.51} & \underline{61.24} & 68.85 & \underline{64.18} \\
            MGCA-Net & \textbf{84.84} & 84.13 & \textbf{83.83} & \textbf{83.62} & 81.07 & \textbf{81.82} & \textbf{74.91} & 74.29 & \textbf{74.31} & \textbf{70.03} & 69.86 & \textbf{69.63} \\
            \bottomrule
        \end{tabular}%
    }
    \caption{\small Quantitative comparison results of the outlier removal task on the YFCC100M and SUN3D datasets are presented as Precision (\%), Recall (\%) and F-score (\%), with the optimal and suboptimal indicators highlighted in bold and underlined, respectively.}
    \label{tab:PRF_result}%
\end{table*}%

\textbf{2) Graph Consensus Construction}:
To capture geometric relationships and enhance the consistency of features across stages, for each stage of feature representation \( \mathbf{Z}, \mathbf{Z}_1^{(M-1)}, \mathbf{Z}_2^{(M-1)}, \mathbf{Z}_3^{(M-1)} \), we separately construct a \( k \)-nearest neighbor graph \( \mathcal{G}_i = (\mathcal{V}_i, \mathcal{E}_i) \) for each feature point , where the set of nodes \( \mathcal{V}_i = \{c_1^i, ... , c_k^i\} \) denotes the feature points and their neighborhoods, the set of edges \( \mathcal{E}_i = \{e_1^i, ... , e_k^i\} \) represents the geometric relationships between feature points. With this sparse graph structure, the geometric correlations between feature points can be dynamically captured, thus providing higher robustness for feature matching in complex scenes.

Furthermore, to improve the consistency of features across stages, we perform high-dimensional feature alignment on the sparse graphs \( \mathcal{G}, \mathcal{G}_1^{(M-1)}, \mathcal{G}_2^{(M-1)}, \mathcal{G}_3^{(M-1)} \) generated at each stage. Specifically, it is completed through the deep concatenation operation of node features \( \mathcal{V} \) and edge features \( \mathcal{E} \), as follows:
\begin{equation}
	\mathcal{G}_{\text{f}} = concat([\mathcal{G}, \mathcal{G}_1^{(M-1)}, \mathcal{G}_2^{(M-1)}, \mathcal{G}_3^{(M-1)}]),
\end{equation}
where \(concat(\cdot)\) represents the concatenation operation, which is used to combine the graph features of different stages into a unified representation space.

Finally, the generated fusion graph \(\mathcal{G}_{\text{f}}\) not only integrates the multi-scale information of features at each stage, but also strengthens the global geometric consistency between nodes and edges through joint representation.

\textbf{3) Cross-stage Graph Consensus Aggregation}:
The cross-stage graph consensus aggregation strategy enhances the global geometric consistency and suppresses the interference of redundant information on subsequent tasks by dynamically weighting and aggregating the feature graphs of each stage at multiple levels. 

First, for the multi-stage information in the consensus feature graph \(\mathcal{G}_{\text{f}}\), we introduce MLP to perform the feature alignment, which aligns the cross-stage feature information and unifies it into a shared feature space, as follows:
\begin{equation} 
	\mathbf{c}_{\text{f}} = MLP\left( \mathcal{G}_{\text{f}} \right),
\end{equation}
where \(\mathbf{c}_{\text{f}}\) is the fused feature graph after MLP processing. 
Then, to aggregate features with geometric consensus relationships of interior points, we employ Annular Convolution (AC) to preserve the relationships between nodes in parallel. Annular Convolution divides \(k\) neighbors into \(k/p\) annular regions based on their affinity to the anchor point and aggregates the contextual information of each annular region using a \(1 \times p\) convolution kernel to maintain the relative relationships between neighbors as follows:
\begin{equation} 
	\tilde{e}_n^i = X W_n e_j^i + b_n, \quad (n-1)p \leq j \leq np,
\end{equation}
where \(\tilde{e}_n^i\) denotes the aggregated features of the \(n\)-th annular region, \(W_n\) and \(b_n\) are learnable parameters, and \(e_j^i\) is the neighbor information of the \(i\)-th feature point. Finally, we process the aggregated features by MLP. 
The strategy ensures that the ring convolutionally aggregated features can fully express local geometric and global semantic information.

\subsection{Loss Function}
\label{sec:loss_function}
In the two-view correspondence task, we aim to classify inliers and outliers while estimating the camera pose via fundamental matrix regression. To optimize both tasks simultaneously, we design a hybrid loss function that combines classification and regression objectives, as follows:
\begin{equation}
	Loss = l_c(W, L) + \gamma l_e(\hat{E}, E),
\end{equation}
where \(l_c(W, L)\) represents the classification loss, which is used to distinguish inliers and outliers; \(\gamma\) is the weight hyperparameter used to balance the classification loss and regression loss, which is set to 0.5, and \(l_e(\hat{E}, E)\) represents the fundamental matrix regression loss, as follows:
\begin{equation}
	l_e(\hat{E}, E) = \frac{(p^T \hat{E} p')^2}{\|Ep\|^2_{[1]} + \|Ep\|^2_{[2]} + \|Ep'\|^2_{[1]} + \|Ep'\|^2_{[2]}},
\end{equation}
where \(\hat{E}\) and \(E\) represent the estimated and true fundamental matrices respectively; \(p\) and \(p'\) are the corresponding feature point coordinates in the two images; the numerator \((p^T \hat{E} p')^2\) represents the squared geometric error of the feature point coordinates under the estimated fundamental matrix; $\Vert \cdot \Vert _{[i]}$ represents the sum of the squares of the elements at the \(i\)-th position in the vector, which ensures the normalization and regularization of the geometric errors at different scales.

\section{Experiments}
\subsection{Implementation Details}
\label{sec:implementation_details}

The number of initial correspondences $N$ for MGCA-Net is set to 2000, with a network dimension of 128. In addition, the number of input neighbors $k$ of each stage of CSMGC is set to 3, and the number of clusters in the Order-Aware block is set to 500.
The training process follows the training strategies from previous benchmarks \cite{zhang2019learning} and CGR-Net \cite{yang2024cgr}, and it is trained for a total of 500k iterations on Ubuntu 18.04 with an NVIDIA GTX 3090.

\begin{table*}[!htbp]
  \centering
  \resizebox{1\linewidth}{!}{
    \begin{tabular}{c|ccccccccccc}
    \toprule
    Dataset & \multicolumn{5}{c}{YFCC100M  (\%)}    &       & \multicolumn{5}{c}{SUN3D  (\%)} \\
    \midrule
    \multirow{2}[4]{*}{Method} & \multicolumn{2}{c}{Known Scene} &       & \multicolumn{2}{c}{Unknown Scene} &       & \multicolumn{2}{c}{Known Scene} &       & \multicolumn{2}{c}{Unknown Scene} \\
\cmidrule{2-3}\cmidrule{5-6}\cmidrule{8-9}\cmidrule{11-12}          & \multicolumn{1}{c|}{5°} & 20°   &       & \multicolumn{1}{c|}{5°} & 20°   &       & \multicolumn{1}{c|}{5°} & 20°   &       & \multicolumn{1}{c|}{5°} & 20° \\
    \midrule
    RANSAC \cite{fischler1981random} & 5.74 /  - & 16.67 /  - &       & 9.05 /  - & 22.71 /  - &       & 4.43 /  - & 15.38 /  - &       & 2.85 /  - & 11.23 /  - \\
    PointNet++ \cite{qi2017pointnet++} & 11.88 /  - & 32.86 /  - &       & 15.98 /  - & 44.82 /  - &       & 8.78 /  - & 31.02 /  - &       & 7.22 /  - & 29.77 /  - \\
    LFGC-Net \cite{yi2018learning} & 14.51 /  - & 35.82 /  - &       & 23.71 /  - & 50.57 /  - &       & 11.93 /  - & 36.03 /  - &       & 9.73 /  - & 33.09 /  - \\
    DFE-Net \cite{ranftl2018deep} & 19.27 /  - & 42.14 /  - &       & 30.55 /  - & 59.15 /  - &       & 14.18 /  - & 39.14 /  - &       & 12.13 /  - & 26.26 /  - \\
    ACNe-Net \cite{sun2020acne} & 29.63 /  - & 52.71 /  - &       & 34.00 /  - & 62.98 /  - &       & 19.08 /  - & 46.32 /  - &       & 14.27 /  - & 39.29 /  - \\
    \midrule
    OANet \cite{zhang2019learning} & 33.75 / 14.49 & 57.13 / 48.04 &       & 40.80 / 17.00 & 69.26 / 58.61 &       & 21.54 / 8.04 & 48.91 / 40.14 &       & 16.37 / 6.09 & 41.82 / 34.08 \\
    T-Net \cite{zhong2021t} & 41.46 / 18.20 & 64.14 / 54.48 &       & 48.40 / 20.70 & 73.92 / 62.94 &       & 23.63 / 9.03 & 51.04 / 42.08 &       & 18.04 / 6.62 & 43.65 / 35.66 \\
    PESA-Net \cite{zhong2022pesa} & 37.15 / 16.09 & 59.76 / 50.32 &       & 45.03 / 19.73 & 71.95 / 61.23 &       & 22.67 / 9.07 & 50.02 / 42.46 &       & 18.00 / 7.26 & 44.10 / 37.47 \\
    MSA-Net \cite{zheng2022msa} & 36.04 / 15.69 & 58.81 / 49.68 &       & 48.70 / 20.95 & 73.07 / 62.31 &       & 19.58 / 7.24 & 47.03 / 38.54 &       & 16.77 / 6.07 & 42.10 / 34.32 \\
    MS²DG-Net \cite{dai2022ms2dg} & 36.46 / 15.36 & 61.94 / 52.18 &       & 46.88 / 18.84 & 74.84 / 63.27 &       & 22.57 / 8.51 & 50.89 / 41.81 &       & 17.19 / 6.24 & 43.27 / 35.22 \\
    U-Match \cite{li2023u} & 46.22 / 21.73 & 67.67 / 57.90 &       & 60.15 / 29.59 & 79.69 / 69.03 &       & 26.45 / 10.20 & 53.56 / 44.38 &       & 22.41 / 8.39 & 48.65 / 40.07 \\
    PGFNet \cite{liu2023pgfnet} & 33.54 / 14.17 & 56.73 / 47.57 &       & 46.70 / 20.48 & 72.26 / 61.62 &       & 22.63 / 8.64 & 49.26 / 40.55 &       & 19.02 / 7.14 & 44.70 / 36.61 \\
    GCA-Net \cite{guo2023learning} & 44.32 / 19.93 & 67.24 / 57.16 &       & 55.70 / 24.90 & 79.58 / 68.21 &       & 22.57 / 8.61 & 50.39 / 41.49 &       & 18.53 / 6.85 & 44.27 / 36.08 \\
    ConvMatch \cite{zhang2023convmatch} & 43.25 / 20.02 & 65.61 / 55.81 &       & 55.45 / 26.69 & 77.53 / 66.87 &       & 27.45 / 10.84 & 54.65 / 45.35 &       & 22.52 / 8.68 & 48.64 / 40.09 \\
    ConvMatch+ \cite{zhang2023convmatch_plus} & 45.79 / 21.19 & 67.69 / 57.72 &       & 58.07 / 27.79 & 78.88 / 68.00 &       & 27.22 / 10.81 & 54.97 / 45.70 &       & 22.67 / 8.62 & 49.13 / 40.51 \\
    NCM-Net \cite{liu2024ncmnet} & 50.24 / 25.02 & 71.27 / 61.31 &       & 62.65 / 32.40 & 82.30 / 71.82 &       & 24.99 / 9.96 & 51.87 / 42.88 &       & 20.41 / 7.92 & 46.42 / 38.16 \\
    PT-Net \cite{gong2024pt} & 49.16 / 20.23 & 71.07 / 56.23 &       & 61.62 / 27.06 & 81.21 / 67.45 &       & 27.23 / 10.67 & 54.38 / 45.11 &       & 22.88 / 8.59 & 48.52 / 39.85 \\
    DeMatch \cite{zhang2024dematch} & 47.56 / 22.51 & 69.14 / 59.17 &       & 60.00 / 30.01 & 80.02 / 69.45 &       & \underline{28.49} / \underline{11.34} & \underline{55.59} / \underline{46.24} &       & \underline{23.46} / \underline{9.20} & \underline{49.84} / \underline{41.20} \\
    BCLNet \cite{miao2024bclnet} & 53.21 / \underline{27.48} & 73.48 / \underline{63.63} &       & \underline{67.85} / \underline{37.22} & \underline{84.57} / \underline{74.44} &       & 24.32 / 9.64 & 51.24 / 42.37 &       & 20.06 / 7.72 & 45.83 / 37.62 \\
    MSGSA \cite{lin2024multi} & \underline{53.87} / 21.99 & \underline{74.04} / 58.26 &       & 63.78 / 28.23 & 82.23 / 68.31 &       & 25.28 / 10.43 & 52.29 / 44.99 &       & 20.41 / 7.81 & 46.12 / 39.00 \\
    CGR-Net \cite{yang2024cgr} & 53.56 / 26.45 & 73.98 / 62.97 &       & 66.47 / 35.01 & 84.14 / 73.17 &       & 26.48 / 10.58 & 53.41 / 44.28 &       & 21.69 / 8.51 & 47.94 / 39.53 \\
    MGCA-Net & \textbf{65.18} / \textbf{35.79} & \textbf{81.40} / \textbf{71.35} &       & \textbf{77.10} / \textbf{44.62} & \textbf{89.23} / \textbf{79.36} &       & \textbf{33.06} / \textbf{14.18} & \textbf{59.22} / \textbf{49.67} &       & \textbf{25.49} / \textbf{10.43} & \textbf{51.54} / \textbf{42.80} \\

    \bottomrule
    \end{tabular}%
    }
    \caption{\small Quantitative comparison results of the relative pose estimation task on the YFCC100M and SUN3D datasets are presented in the form of mAP \ AUC at 5° and 20°, with the best and second-best results highlighted in bold and underlined, respectively.}
  \label{tab:mAP_AUC_combine_result}%
\end{table*}%

\subsection{Datasets and Evaluation Metrics}
To evaluate the proposed MGCA-Net, we use representative outdoor and indoor datasets for training and testing.

\textbf{Outdoor Dataset:} The YFCC100M dataset~\cite{thomee2016yfcc100m} contains 99.2 million images and 0.8 million videos with rich metadata. We select 72 outdoor scene sequences, with 68 used for training, validation, and testing, and 4 as unknown scenes for generalization evaluation.

\textbf{Indoor Dataset:} The SUN3D dataset~\cite{xiao2013sun3d} consists of 254 RGB-D video scenes. We use 239 for training, validation, and testing, and 15 for evaluating generalization.

\textbf{Outlier Rejection:} 
\label{sec:outlier_rejection}
To evaluate the effectiveness of the proposed MGCA-Net in the outlier rejection task, we used three common metrics: Precision (P), Recall (R), and F-score (F), which reflect the effectiveness of MGCA-Net in outlier identification and removal.

\textbf{Camera Pose Estimation:} 
For camera pose estimation task, we assess the performance of MGCA-Net using Mean Average Precision (mAP) and Area Under the Curve (AUC) at 5$^\circ$ and 20$^\circ$ thresholds.

\subsection{Outlier Rejection}
To evaluate the effectiveness of MGCA-Net, we compare it with 21 representative methods, including traditional approaches such as RANSAC~\cite{fischler1981random}; MLP-based deep learning methods such as PointNet++~\cite{qi2017pointnet++}, LFGC-Net~\cite{yi2018learning}, OANet~\cite{zhang2019learning}, NCM-Net~\cite{liu2024ncmnet}, and CGR-Net~\cite{yang2024cgr}; and CNN-based methods such as ConvMatch\textsuperscript{+}~\cite{zhang2023convmatch_plus}, PT-Net~\cite{gong2024pt}, and DeMatch~\cite{zhang2024dematch}. Experimental results for RANSAC~\cite{fischler1981random}, PointNet++~\cite{qi2017pointnet++}, LFGC-Net~\cite{yi2018learning}, DFE-Net~\cite{ranftl2018deep}, and ACNe-Net~\cite{sun2020acne} are cited from T-Net~\cite{zhong2021t}, while other results are obtained using publicly available code under consistent experimental settings for fair comparison.

As reported in Table \ref{tab:PRF_result}, MGCA-Net outperforms other competing methods on both the YFCC and SUN3D datasets, demonstrating its significant advantage in the two-view outlier removal task.
For the YFCC100M dataset, in known scenes, MGCA-Net's P and F scores reach 84.84\% and 83.83\%, respectively, which are 6.48\% and 3.93\% higher than those of the second-ranked BCLNet. In unknown scenes, MGCA-Net's P and F scores are respectively 5.72\% and 3.09\% higher than those of BCLNet.
For the SUN3D dataset, in known scenes, MGCA-Net's P and F scores are 8.45\% and 4.8\% higher than those of the second-ranked CGR-Net, while in unknown scenes, they achieve improvements of 8.79\% and 5.45\%. Experimental results demonstrate that MGCA-Net achieves higher accuracy and robustness than existing methods across different scenes.

\begin{figure}[!t]
    \centering
    \includegraphics[width=1\linewidth]{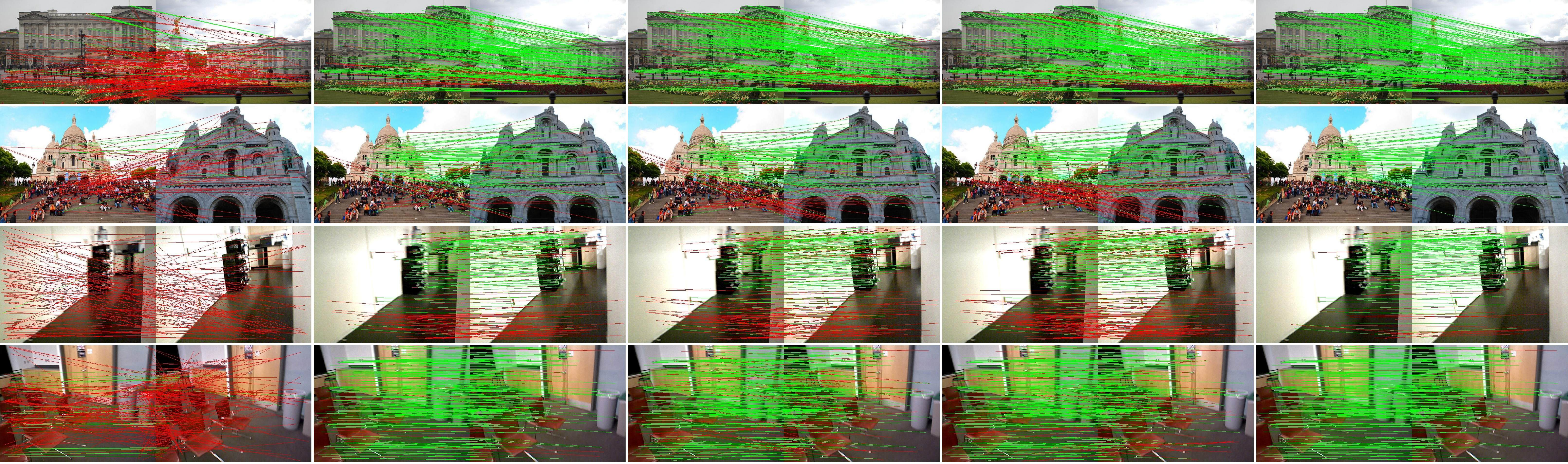} 
    \vspace{-6pt}
    \makebox[\linewidth]{
            \small
            (a) BCLNet \hspace{0.1cm}
            (b) MSGSA 
            (c) ConvMatch 
            (d) PT-Net 
            (e) MGCA-Net
        }
    \caption{Qualitative results of outlier removal. The first and second rows show outdoor scenes from YFCC100M, while the third and fourth rows depict indoor scenes from SUN3D. False matches are marked in red and correct matches are marked in green.}
    \label{fig:match_vis}
\end{figure}

Although MGCA-Net achieves high Precision and F-score, its Recall is slightly lower than DeMatch and ConvMatch in some cases, mainly due to different logit threshold settings. Lower thresholds improve Recall but reduce Precision and F-score. As illustrated in Fig.~\ref{fig:match_vis}, MGCA-Net yields fewer incorrect matches and superior overall matching in both indoor and outdoor scenes, highlighting its robustness and accuracy.

\subsection{Camera Pose Estimation}
To assess the performance of MGCA-Net in camera pose estimation, we compared MGCA-Net with 16 representative methods, including traditional approaches such as RANSAC \cite{fischler1981random}; MLP-based deep learning methods such as PointNet++ \cite{qi2017pointnet++}, LFGC-Net \cite{yi2018learning}, DFE-Net \cite{ranftl2018deep}, and ACNe-Net \cite{sun2020acne}; and CNN-based methods. The results for RANSAC, PointNet++, LFGC-Net, DFE-Net, and ACNe-Net are cited from T-Net \cite{zhong2021t}. 

As reported in Table~\ref{tab:mAP_AUC_combine_result}, MGCA-Net achieves substantial gains over the second-best methods on YFCC100M and SUN3D, covering both indoor and outdoor scenes. On YFCC100M, MGCA-Net outperforms traditional methods~\cite{fischler1981random} in mAP@5° and mAP@20° by 59.44\% and 64.73\% for known scenes, and by 68.05\% and 66.52\% for unknown scenes. On SUN3D, the improvements reach 28.63\% and 43.84\% (known), and 22.64\% and 40.31\% (unknown). Compared with SOTA deep learning approaches such as MSGSA~\cite{lin2024multi}, BCLNet~\cite{miao2024bclnet}, and DeMatch~\cite{zhang2024dematch}, MGCA-Net achieves higher mAP@5°, mAP@20°, AUC@5°, and AUC@20° by 11.31\%, 7.36\%, 8.31\%, and 7.72\% in known YFCC100M scenes, and by 9.25\%, 4.66\%, 7.4\%, and 4.92\% in unknown scenes; for SUN3D, the gains are 4.57\%, 3.63\%, 2.84\%, and 3.43\% (known), and 2.03\%, 1.7\%, 1.23\%, and 1.6\% (unknown), respectively.

The above experimental results show that MGCA-Net performs well in different datasets and scenes, especially in complex scenes. This is due to the fact that MGCA-Net effectively combines the feature representations of content and location through CGA, which in turn improves the feature extraction capability for local geometric consistency and global contextual information. Secondly, based on CSMGC, MGCA-Net is able to integrate multi-scale geometric features across stages and construct cross-stage geometric consensus relations through multiple different sparse graphs.

\subsection{Ablation Experiments}
\label{sec:Ablation_Experiments}
MGCA-Net consists of two main modules: CGA, which serves as the core for contextual and geometric feature extraction, and CSMGC, which enhances cross-stage geometric consistency. To evaluate their effectiveness, we conducted ablation experiments on unknown scenes from YFCC100M, as shown in Table~\ref{tab:ablation_experiments}. The results demonstrate that each module incrementally improves performance, each component contributes to the network performance as follows:

\textbf{Baseline:} When only the basic module is used for feature extraction, mAP@5° and mAP@20° are 56.70\% and 79.33\%, respectively, highlighting the limitations without contextual and geometric consistency information.

\textbf{Iter:} After adding the multi-stage iterative structure, mAP5° and mAP20° reach 58.20\% and 80.46\%, respectively, demonstrating that iterative optimization improves feature extraction, though performance remains limited without effective use of intermediate information.

\textbf{CGA:} After adding CGA, the network performance is further improved, and mAP5° and mAP20° reach 66.57\% and 84.58\%, respectively, which shows that the fusion of context features and graph coordinate features contributes significantly to the model performance.

\textbf{Iter + CGA:} After combining the iterative network and CGA module, mAP5° and mAP20° reach 69.88\% and 86.39\% respectively, which reflects the importance of the collaboration between them.

\textbf{Iter + CSMGC:} After combining the iterative network and CSMGC, mAP5° and mAP20° reach 73.70\% and 88.26\% respectively, which verifies the importance of cross-stage feature fusion and geometric consistency modeling.

\textbf{Full Model:} With CGA, CSMGC, and multi-stage iteration, mAP@5° and mAP@20° reach 77.10\% and 89.23\%, significantly outperforming other settings. These results confirm that combining CGA and CSMGC with multi-stage structure fully unlocks MGCA-Net's potential for geometric feature representation and matching.

\begin{table}[!t]
    \centering
    \resizebox{0.8\linewidth}{!}{
        \begin{tabular}{cccc|cc}
            \toprule
            Baseline & Iter  & CGA   & CSMGC & mAP5° (\%) & mAP20° (\%) \\
            \midrule
            \checkmark &       &       &       & 56.70 & 79.33 \\
            \checkmark &       & \checkmark &       & 58.20 & 80.46 \\
            \checkmark & \checkmark &       &       & 66.57 & 84.58 \\
            \checkmark & \checkmark & \checkmark &       & 69.88 & 86.39 \\
            \checkmark & \checkmark &       & \checkmark & 73.70 & 88.26 \\
            \checkmark & \checkmark & \checkmark & \checkmark & 77.10 & 89.23 \\
            \bottomrule
        \end{tabular}%
    }
    \caption{\small Ablation study on unknown scenes with different modules.}
    \label{tab:ablation_experiments}
\end{table}%
\section{Conclusion}
In this paper, we propose MGCA-Net, a simple yet effective framework for two-view geometric correspondence learning that captures deep geometric relationships by integrating contextual and positional information. MGCA-Net comprises the CGA and CSMGC modules, which jointly enhance feature representation and establish geometric consensus across different stages. Experimental results on the YFCC100M and SUN3D datasets demonstrate that MGCA-Net achieves robust performance in both indoor and outdoor scenes, showing stable outlier rejection and significant improvements in camera pose estimation. Compared to traditional and state-of-the-art deep learning methods, MGCA-Net exhibits superior adaptability in terms of matching accuracy and robustness, particularly in complex scenarios. In future work, we will explore extending MGCA-Net to unsupervised or self-supervised learning frameworks to reduce reliance on labeled data and enhance its ability to generalize in different domains.


	%
	%
	%
    
	\section*{Acknowledgments}
	This work was supported in part by the National Natural Science Foundation of China (Grant Nos. U22A2095, 62476112, 62202249); in part by the Guangdong Basic and Applied Basic Research Foundation (Grant Nos. 2024A1515011740, 2025A1515010181); in part by the Fundamental Research Funds for the Central Universities (Grant Nos. 21624404, 23JNSYS01); in part by the China Postdoctoral Science Foundation (Grant Nos. GZC20233362, 2024MD754043) and Chongqing Municipal Education Commission (Grant No. KJQN202400648); in part by the Major Key Project of PCL (Grant No. PCL2024A04-4); in part by Guangdong Key Laboratory of Data Security and Privacy Preserving (Grant No. 2023B1212060036); and in part by Guangdong-Hong Kong Joint Laboratory for Data Security and Privacy Preserving (Grant No. 2023B1212120007).

	\bibliographystyle{named}
	\bibliography{Ref}

@String(AAAI = {AAAI})

@inproceedings{schonberger2016structure,
  title={Structure-from-motion revisited},
  author={Schonberger, Johannes L and Frahm, Jan-Michael},
  booktitle={Proceedings of the IEEE Conference on Computer Vision and Pattern Recognition},
  pages={4104--4113},
  year={2016}
}

@article{fischler1981random,
  title={Random sample consensus: a paradigm for model fitting with applications to image analysis and automated cartography},
  author={Fischler, Martin A and Bolles, Robert C},
  journal={Communications of the ACM},
  volume={24},
  number={6},
  pages={381--395},
  year={1981},
  publisher={ACM New York, NY, USA}
}

@Article{gong2024pt,
  author    = {Gong, Zhepeng and Xiao, Guobao and Shi, Ziwei and Wang, Shiping and Chen, Riqing},
  journal   = {IEEE Transactions on Instrumentation and Measurement},
  title     = {PT-Net: Pyramid transformer network for feature matching learning},
  year      = {2024},
  pages     = {1--11},
  volume    = {73},
  publisher = {IEEE},
}

@Article{zhang2023convmatch_plus,
  author    = {Zhang, Shihua and Ma, Jiayi},
  journal   = {IEEE Transactions on Pattern Analysis and Machine Intelligence},
  title     = {Convmatch: Rethinking network design for two-view correspondence learning},
  year      = {2023},
  number    = {5},
  pages     = {2920--2935},
  volume    = {46},
  publisher = {IEEE},
}

@article{ma2021image,
  title={Image matching from handcrafted to deep features: A survey},
  author={Ma, Jiayi and Jiang, Xingyu and Fan, Aoxiang and Jiang, Junjun and Yan, Junchi},
  journal={International Journal of Computer Vision},
  pages={23--79},
  year={2021}
}

@article{lowe2004distinctive,
  title={Distinctive image features from scale-invariant keypoints},
  author={Lowe, David G},
  journal={International Journal of Computer Vision},
  pages={91--110},
  year={2004}
}

@inproceedings{detone2018superpoint,
  title={Superpoint: Self-supervised interest point detection and description},
  author={DeTone, Daniel and Malisiewicz, Tomasz and Rabinovich, Andrew},
  booktitle={Proceedings of the IEEE Conference on Computer Vision and Pattern Recognition},
  pages={224--236},
  year={2018}
}

@article{ma2014robust,
  title={Robust point matching via vector field consensus},
  author={Ma, Jiayi and Zhao, Ji and Tian, Jinwen and Yuille, Alan L and Tu, Zhuowen},
  journal={IEEE Transactions on Image Processing},
  pages={1706--1721},
  year={2014}
}

@inproceedings{yi2018learning,
  title={Learning to find good correspondences},
  author={Yi, Kwang Moo and Trulls, Eduard and Ono, Yuki and Lepetit, Vincent and Salzmann, Mathieu and Fua, Pascal},
  booktitle={Proceedings of the IEEE Conference on Computer Vision and Pattern Recognition},
  pages={2666--2674},
  year={2018}
}

@inproceedings{sun2020acne,
  title={Acne: Attentive context normalization for robust permutation-equivariant learning},
  author={Sun, Weiwei and Jiang, Wei and Trulls, Eduard and Tagliasacchi, Andrea and Yi, Kwang Moo},
  booktitle={Proceedings of the IEEE Conference on Computer Vision and Pattern Recognition},
  pages={11286--11295},
  year={2020}
}

@inproceedings{zhang2019learning,
  title={Learning two-view correspondences and geometry using order-aware network},
  author={Zhang, Jiahui and Sun, Dawei and Luo, Zixin and Yao, Anbang and Zhou, Lei and Shen, Tianwei and Chen, Yurong and Quan, Long and Liao, Hongen},
  booktitle={Proceedings of the IEEE International Conference on Computer Vision},
  pages={5845--5854},
  year={2019}
}

@inproceedings{zhong2021t,
  title={T-Net: Effective permutation-equivariant network for two-view correspondence learning},
  author={Zhong, Zhen and Xiao, Guobao and Zheng, Linxin and Lu, Yan and Ma, Jiayi},
  booktitle={Proceedings of the IEEE International Conference on Computer Vision},
  pages={1950--1959},
  year={2021}
}

@Article{zhong2022pesa,
  author  = {Zhong, Zhen and Xiao, Guobao and Wang, Shiping and Wei, Leyi and Zhang, Xiaoqin},
  journal = {Information Fusion},
  title   = {Pesa-net: Permutation-equivariant split attention network for correspondence learning},
  year    = {2022},
  pages   = {81--89},
}

@Article{zheng2022msa,
  author  = {Zheng, Linxin and Xiao, Guobao and Shi, Ziwei and Wang, Shiping and Ma, Jiayi},
  journal = {IEEE Transactions on Image Processing},
  title   = {MSA-Net: Establishing reliable correspondences by multiscale attention network},
  year    = {2022},
  pages   = {4598--4608},

}

@inproceedings{dai2022ms2dg,
  title={MS2DG-Net: Progressive correspondence learning via multiple sparse semantics dynamic graph},
  author={Dai, Luanyuan and Liu, Yizhang and Ma, Jiayi and Wei, Lifang and Lai, Taotao and Yang, Changcai and Chen, Riqing},
  booktitle={Proceedings of the IEEE Conference on Computer Vision and Pattern Recognition},
  pages={8973--8982},
  year={2022}
}

@article{raguram2012usac,
  title={USAC: A universal framework for random sample consensus},
  author={Raguram, Rahul and Chum, Ondrej and Pollefeys, Marc and Matas, Jiri and Frahm, Jan-Michael},
  journal={IEEE Transactions on Pattern Analysis and Machine Intelligence},
  pages={2022--2038},
  year={2012}
}

@Article{qi2017pointnet++,
  author  = {Qi, Charles Ruizhongtai and Yi, Li and Su, Hao and Guibas, Leonidas J},
  journal = {Proceedings of the Advances in Neural Information Processing Systems},
  title   = {Pointnet++: Deep hierarchical feature learning on point sets in a metric space},
  year    = {2017},
  pages   = {5099--5108},
}

@inproceedings{ranftl2018deep,
  title={Deep fundamental matrix estimation},
  author={Ranftl, Ren{\'e} and Koltun, Vladlen},
  booktitle={Proceedings of the European Conference on Computer Vision},
  pages={284--299},
  year={2018}
}

@Article{lin2022co,
  author  = {Lin, Shuyuan and Luo, Hailing and Yan, Yan and Xiao, Guobao and Wang, Hanzi},
  journal = {IEEE Transactions on Image Processing},
  title   = {Co-clustering on bipartite graphs for robust model fitting},
  year    = {2022},
  volume = {31},
  pages   = {6605--6620},
}

@article{thomee2016yfcc100m,
  title={YFCC100M: The new data in multimedia research},
  author={Thomee, Bart and Shamma, David A and Friedland, Gerald and Elizalde, Benjamin and Ni, Karl and Poland, Douglas and Borth, Damian and Li, Li-Jia},
  journal={Communications of the ACM},
  pages={64--73},
  year={2016}
}

@inproceedings{xiao2013sun3d,
  title={Sun3d: A database of big spaces reconstructed using sfm and object labels},
  author={Xiao, Jianxiong and Owens, Andrew and Torralba, Antonio},
  booktitle={Proceedings of the IEEE International Conference on Computer Vision},
  pages={1625--1632},
  year={2013}
}

@Article{liao2023sga,
  author  = {Liao, Tangfei and Zhang, Xiaoqin and Xu, Yuewang and Shi, Ziwei and Xiao, Guobao},
  journal = {IEEE Transactions on Circuits and Systems for Video Technology},
  title   = {SGA-Net: A sparse graph attention network for two-view correspondence learning},
  year    = {2023},
  number  = {12},
  pages   = {7578--7590},
  volume  = {33},
}

@article{liu2023pgfnet,
  title={Pgfnet: Preference-guided filtering network for two-view correspondence learning},
  author={Liu, Xin and Xiao, Guobao and Chen, Riqing and Ma, Jiayi},
  journal={IEEE Transactions on Image Processing},
  volume={32},
  pages={1367--1378},
  year={2023},
  publisher={IEEE}
}

@article{guo2023learning,
  title={Learning for feature matching via graph context attention},
  author={Guo, Junwen and Xiao, Guobao and Tang, Zhimin and Chen, Shunxing and Wang, Shiping and Ma, Jiayi},
  journal={IEEE Transactions on Geoscience and Remote Sensing},
  volume={61},
  pages={1--14},
  year={2023},
  publisher={IEEE}
}

@Article{liu2024ncmnet,
  author    = {Liu, Xin and Qin, Rong and Yan, Junchi and Yang, Jufeng},
  journal   = {IEEE Transactions on Pattern Analysis and Machine Intelligence},
  title     = {NCMNet: Neighbor consistency mining network for two-view correspondence pruning},
  year      = {2024},
  number    = {12},
  pages     = {11254--11272},
  volume    = {46},
  publisher = {IEEE},
}

@inproceedings{miao2024bclnet,
  title={Bclnet: Bilateral consensus learning for two-view correspondence pruning},
  author={Miao, Xiangyang and Xiao, Guobao and Wang, Shiping and Yu, Jun},
  booktitle={Proceedings of the AAAI Conference on Artificial Intelligence},
  volume={38},
  number={5},
  pages={4225--4232},
  year={2024}
}

@Article{lin2024multi,
  author    = {Lin, Shuyuan and Chen, Xiao and Xiao, Guobao and Wang, Hanzi and Huang, Feiran and Weng, Jian},
  journal   = {IEEE Transactions on Image Processing},
  title     = {Multi-Stage Network with Geometric Semantic Attention for Two-View Correspondence Learning},
  year      = {2024},
  pages     = {3031--3046},
  volume    = {33},
  publisher = {IEEE},
}

@Article{yang2024cgr,
  author    = {Yang, Changcai and Li, Xiaojie and Ma, Jiayi and Zhuang, Fengyuan and Wei, Lifang and Chen, Riqing and Chen, Guodong},
  journal   = {IEEE Transactions on Circuits and Systems for Video Technology},
  title     = {CGR-Net: Consistency guided resformer for two-view correspondence learning},
  year      = {2024},
  number    = {12},
  pages     = {12450--12465},
  volume    = {34},
  publisher = {IEEE},
}

@InProceedings{sarlin2020superglue,
  author    = {Sarlin, Paul-Edouard and DeTone, Daniel and Malisiewicz, Tomasz and Rabinovich, Andrew},
  booktitle = {Proceedings of the IEEE Conference on Computer Vision and Pattern Recognition},
  title     = {Superglue: Learning feature matching with graph neural networks},
  year      = {2020},
  pages     = {4938--4947},
}

@InProceedings{sun2021loftr,
  author    = {Sun, Jiaming and Shen, Zehong and Wang, Yuang and Bao, Hujun and Zhou, Xiaowei},
  booktitle = {Proceedings of the IEEE Conference on Computer Vision and Pattern Recognition},
  title     = {LoFTR: Detector-free local feature matching with transformers},
  year      = {2021},
  pages     = {8922--8931},
}

@InProceedings{li2023u,
  author    = {Li, Zizhuo and Zhang, Shihua and Ma, Jiayi},
  booktitle = {Proceedings of the International Joint Conference on Artificial Intelligence},
  title     = {U-Match: Two-view correspondence learning with hierarchy-aware local context aggregation.},
  year      = {2023},
  pages     = {1169--1176},
}

@InProceedings{zhang2024dematch,
  author    = {Zhang, Shihua and Li, Zizhuo and Gao, Yuan and Ma, Jiayi},
  booktitle = {Proceedings of the IEEE Conference on Computer Vision and Pattern Recognition},
  title     = {DeMatch: Deep decomposition of motion field for two-view correspondence learning},
  year      = {2024},
  pages     = {20278--20287},
}

@Article{chen2024location,
  author   = {Chen, Changhao and Wang, Bing and Lu, Chris Xiaoxuan and Trigoni, Niki and Markham, Andrew},
  journal  = {IEEE Transactions on Neural Networks and Learning Systems},
  title    = {Deep learning for visual localization and mapping: A survey},
  year     = {2024},
  number   = {12},
  pages    = {17000-17020},
  volume   = {35},
  doi      = {10.1109/TNNLS.2023.3309809},
}

@article{placed2023survey,
  title={A survey on active simultaneous localization and mapping: State of the art and new frontiers},
  author={Placed, Julio A and Strader, Jared and Carrillo, Henry and Atanasov, Nikolay and Indelman, Vadim and Carlone, Luca and Castellanos, Jos{\'e} A},
  journal={IEEE Transactions on Robotics},
  volume={39},
  number={3},
  pages={1686--1705},
  year={2023},
  publisher={IEEE}
}

@InProceedings{schmied2023r3d3,
  author    = {Schmied, Aron and Fischer, Tobias and Danelljan, Martin and Pollefeys, Marc and Yu, Fisher},
  booktitle = {Proceedings of the IEEE International Conference on Computer Vision},
  title     = {R3d3: Dense 3d reconstruction of dynamic scenes from multiple cameras},
  year      = {2023},
  pages     = {3216--3226},
}

@InProceedings{brachmann2019neural,
  author    = {Brachmann, Eric and Rother, Carsten},
  booktitle = {Proceedings of the IEEE International Conference on Computer Vision},
  title     = {Neural-guided RANSAC: Learning where to sample model hypotheses},
  year      = {2019},
  pages     = {4322--4331},
}

@inproceedings{barath2018graph,
  title={Graph-cut RANSAC},
  author={Barath, Daniel and Matas, Ji{\v{r}}{\'\i}},
  booktitle={Proceedings of the IEEE Conference on Computer Vision and Pattern Recognition},
  pages={6733--6741},
  year={2018}
}

@InProceedings{zhao2021progressive,
  author    = {Zhao, Chen and Ge, Yixiao and Zhu, Feng and Zhao, Rui and Li, Hongsheng and Salzmann, Mathieu},
  booktitle = {Proceedings of the IEEE International Conference on Computer Vision},
  title     = {Progressive correspondence pruning by consensus learning},
  year      = {2021},
  pages     = {6464--6473},
}

@InProceedings{liao2024vsformer,
  author    = {Liao, Tangfei and Zhang, Xiaoqin and Zhao, Li and Wang, Tao and Xiao, Guobao},
  booktitle = {Proceedings of the AAAI Conference on Artificial Intelligence},
  title     = {VSFormer: Visual-spatial fusion transformer for correspondence pruning},
  year      = {2024},
  number    = {4},
  pages     = {3369--3377},
  volume    = {38},
}

@article{ma2019locality,
  title={Locality preserving matching},
  author={Ma, Jiayi and Zhao, Ji and Jiang, Junjun and Zhou, Huabing and Guo, Xiaojie},
  journal={International Journal of Computer Vision},
  volume={127},
  pages={512--531},
  year={2019},
  publisher={Springer}
}

@InProceedings{zhang2023convmatch,
  author    = {Zhang, Shihua and Ma, Jiayi},
  booktitle = {Proceedings of the AAAI Conference on Artificial Intelligence},
  title     = {ConvMatch: Rethinking network design for two-view correspondence learning},
  year      = {2023},
  number    = {3},
  pages     = {3472--3479},
  volume    = {37},
}

@article{lin2024robust,
  title = {Robust heterogeneous model fitting for multi-source image correspondences},
  author = {Lin, Shuyuan and Huang, Feiran and Lai, Taotao and Lai, Jianhuang and Wang, Hanzi and Weng, Jian},
  year = {2024},
  journal = {International Journal of Computer Vision},
  volume = {132},
  pages = {2907--2928},
}
	
\end{document}